\documentclass[letterpaper, 10 pt, conference]{ieeeconf}  

\IEEEoverridecommandlockouts                              

\usepackage{amsmath,amsfonts}
\usepackage[algo2e,ruled,vlined,linesnumbered,resetcount,norelsize]{algorithm2e}
\usepackage{array}
\usepackage{textcomp}
\usepackage{stfloats}
\usepackage{float}
\usepackage{url}
\usepackage{verbatim}
\usepackage{graphicx}
\usepackage{cite}
\usepackage{mathtools} 
\usepackage{amssymb}  

\usepackage{amsthm}
\usepackage{algpseudocode}
\usepackage{xcolor}
\usepackage{accents}
\usepackage{mathrsfs}
\usepackage[normalem]{ulem}

\def\beq{\begin{equation*}}
\def\eeq{\end{equation*}}
\def\bql{\begin{equation}}
\def\eql{\end{equation}}
\def\bqn{\begin{eqnarray*}}
\def\eqn{\end{eqnarray*}}
\def\bnl{\begin{eqnarray}}
\def\enl{\end{eqnarray}}

\usepackage{soul}
\usepackage{booktabs}
\usepackage{amsmath} 

\usepackage{multirow}

\usepackage[T1]{fontenc}
\usepackage[utf8]{inputenc}
\usepackage[skip=0.333\baselineskip]{caption}
\usepackage{siunitx} 
\newcolumntype{T}{S[table-format=3.3, input-symbols={()},
                    table-space-text-post={$^{***}$},
                    table-align-text-post=false]}
\usepackage[colorlinks=true, allcolors=black]{hyperref}
\usepackage{tabularx,booktabs}
\newcolumntype{C}{>{\centering\arraybackslash}X} 
\definecolor{green}{RGB}{11,155,13}

\SetCommentSty{mycommfont}

\SetKwRepeat{Do}{do}{while}

\DeclareMathOperator*{\argmax}{argmax}

\usepackage{yaacro}

\begin{acgroupdef}[list=acronimos]
    \acdef{JSG}{Joint State Graph}
\end{acgroupdef}

\newcommand{\nodeset}{\mathbb{V}}
\newcommand{\edgeset}{\mathbb{E}}
\newcommand{\agentset}{\mathbb{N}}
\newcommand{\edge}[2]{e_{#1,#2}} 

\newcommand{\supnodes}[2]{\mathcal{Z}_{#1,#2}}
\newcommand{\cost}[2]{c_{#1,#2}} 
\newcommand{\supcost}{\tilde{c}} 
\newcommand{\supmove}[2]{\tilde{c}_{#1,#2}} 

\newcommand{\actset}[1]{A_{#1}}
\newcommand{\agentcost}[2]{c^{#1}_{#2}}
\newcommand{\actseq}[1]{\mathcal{A}^{#1}}
\newcommand{\nodeseq}[1]{\mathcal{V}^{#1}}
\newcommand{\act}[3]{a^{#1}_{#2 #3}}

\title{\LARGE \bf
Scaling Team Coordination on Graphs with Reinforcement Learning
}

\author{Manshi Limbu, Zechen Hu, Xuan Wang, Daigo Shishika$^\dagger$, and Xuesu Xiao$^\dagger$
\thanks{$^\dagger$Equally advising authors, George Mason University. {\tt\scriptsize \{klimbu2, zhu3, xwang64, dshishik, xiao\}@gmu.edu}. 
Work supported by ARL Tactical Behaviors for Autonomous Maneuver (TBAM) CRA.
This work has taken place in the RobotiXX Laboratory at GMU. RobotiXX research is supported by ARO (W911NF2220242, W911NF2320004, W911NF2420027), AFCENT, Google DeepMind, Clearpath Robotics, and Raytheon Technologies. 
}
}

\begin{document}
\maketitle
\thispagestyle{empty}
\pagestyle{empty}


\begin{abstract}
    This paper studies Reinforcement Learning (RL) techniques to enable team coordination behaviors in graph environments with support actions among teammates to reduce the costs of traversing certain risky edges in a centralized manner. 
    While classical approaches can solve this non-standard multi-agent path planning problem by converting the original Environment Graph (\textsc{eg}) into a Joint State Graph (\textsc{jsg}) to implicitly incorporate the support actions, those methods do not scale well to large graphs and teams. 
    To address this curse of dimensionality, we propose to use RL to enable agents to \emph{learn} such graph traversal and teammate supporting behaviors in a data-driven manner. 
    Specifically, through a new formulation of the team coordination on graphs with risky edges problem into Markov Decision Processes (MDPs) with a novel state and action space, we investigate how RL can solve it in two paradigms: 
    First, we use RL for a team of agents to learn how to coordinate and reach the goal with minimal cost on a single \textsc{eg}. We show that RL efficiently solves problems with up to 20/4 or 25/3 nodes/agents, using a fraction of the time needed for \textsc{jsg} to solve such complex problems; 
    Second, we learn a general RL policy for any $N$-node \textsc{eg}s to produce efficient supporting behaviors.  
    We present extensive experiments and compare our RL approaches against their classical counterparts. 
\end{abstract}

\section{INTRODUCTION}

Multi-robot systems have been studied with the premise of increased efficacy using many low-capability robots as opposed to a small number of high-capability robots. In such a setting, the coordination between low-capability teammates is essential to achieve the whole team's high efficacy. 
This paper is interested in a scenario where a team of robots cooperatively traverse a challenging environment by ``supporting'' each other. Support can take the form of, for example, providing a different vantage point for better situational awareness, or physically interacting with the environment to reduce risk (e.g., holding a ladder). We abstract these notions to the actions that can be taken on a graph environment to study how multi-robot teams can efficiently traverse an environment when such cooperation is possible.

Team coordination on graphs with state-dependent edge cost~\cite{limbu2023team} is a recently proposed problem, in which a team of agents  move on an Environment Graph (\textsc{eg}) and provide support actions for teammates to reduce the cost to traverse certain risky edges with the goal of achieving minimal traversal cost for the whole team to reach the goal(s). Prior methods convert the \textsc{eg} with risky edges (whose traversal cost depends on whether a teammate is supporting the traversal from a certain support node) into a Joint State Graph (\textsc{jsg}) which implicitly incorporates the support actions and then applies graph-search algorithms to find the minimal cost path on the \textsc{jsg}. However, \textsc{jsg} does not scale well to larger graphs and team sizes, while its extended version Critical Joint State Graph (\textsc{cjsg}) can only efficiently handle two agents on graphs where the ratio between risky edge and normal edge is low.

\begin{figure}[t]
    \centering
    \includegraphics[width=0.235\textwidth]{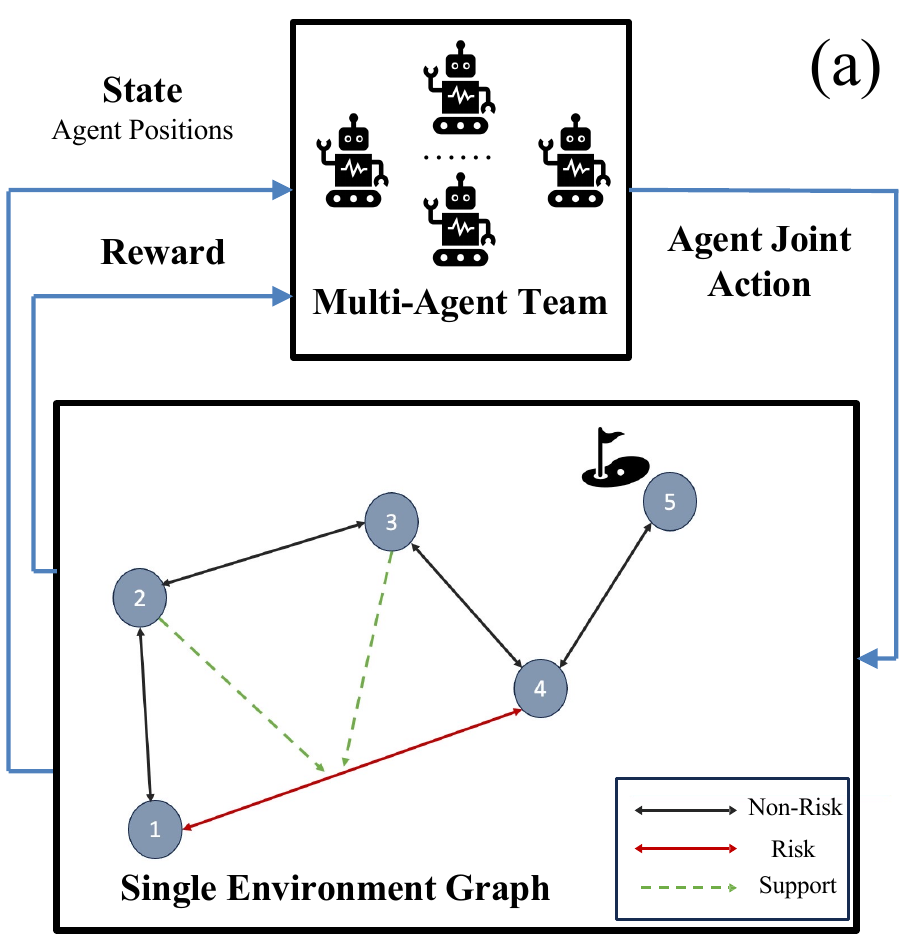}
    \includegraphics[width=0.235\textwidth]{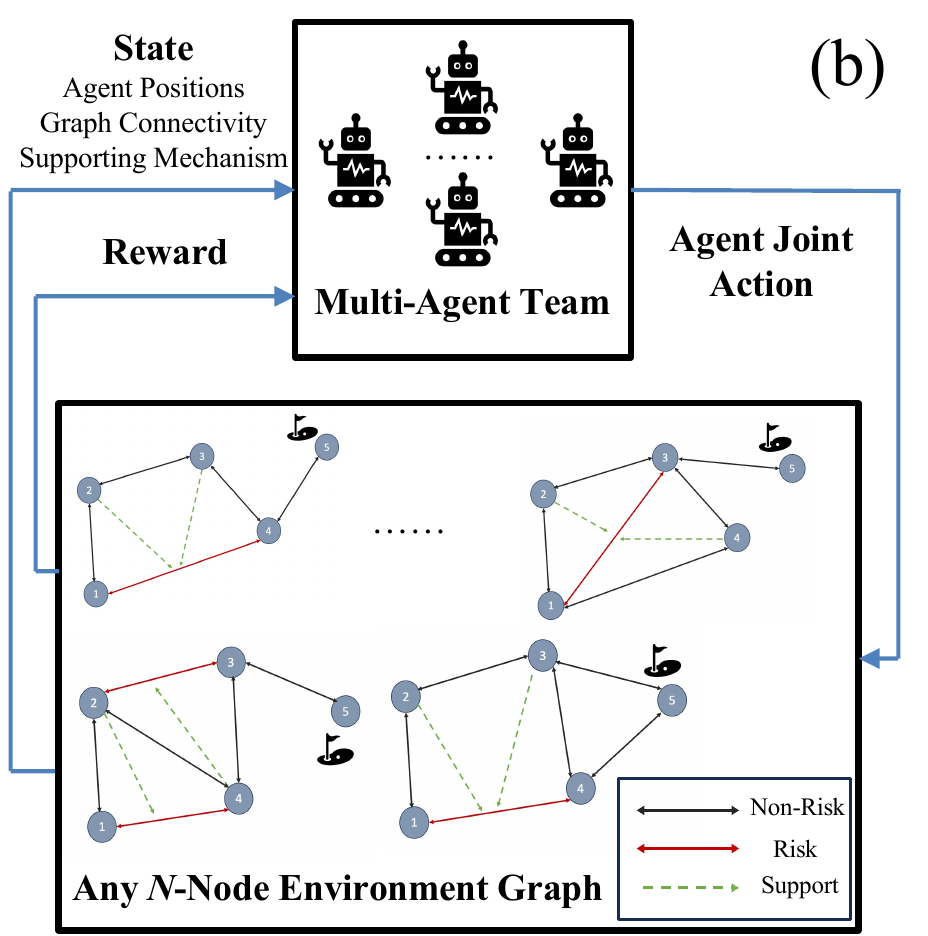}
    \caption{Team coordination with reinforcement learning on a single graph (a) and on multiple graphs (b) with risky edges and supporting behaviors to reduce risk.}
    \label{fig::fig1_problem}
    \vspace{-20pt}
\end{figure}

Reinforcement Learning (RL) has the potential to allow agents to learn from trial-and-error experiences by exploring on the graph. It has the potential to generate coordination behaviors on the \textsc{eg} without constructing \textsc{jsg} or \textsc{cjsg} and searching on those large-scale, densely connected graphs. In this paper, we investigate how the original centralized problem of team coordination on graphs with risky edges can be converted into Markov Decision Processes (MDPs) with graph-dependent state and action spaces and then be solved by state-of-the-art RL techniques. To be specific, we are interested in using RL to solve the team coordination problem in two paradigms. In the first paradigm (Fig.~\ref{fig::fig1_problem} (a)), RL is used to solve the team coordination on one single \textsc{eg} with more than two agents. Our experiment results show that the total time including training and inference of the RL policy can be faster than the time used by the \textsc{jsg} approach, including \textsc{jsg} construction and search time, in complex problems with many agents and nodes. RL can also extend to more than two agents, which \textsc{cjsg} cannot solve; In the second paradigm (Fig.~\ref{fig::fig1_problem} (b)), we use RL to solve any $N$-node graph with a novel formulation of the agent state, graph connectivity, and supporting mechanism encoded in the state space. Our experiment results show that a general RL policy can be learned for any graph and support structure up to ten nodes to produce efficient supporting behaviors. To be best of our knowledge, this work is the first to utilize RL to solve a multi-robot coordination problem on \emph{graphs}. 
\section{RELATED WORK}
\label{sec::related}
We review related work on multi-agent systems and multi-agent reinforcement learning techniques. 

\subsection{Multi-Agent Systems}
Multi-agent systems have received tremendous attention from different fields, ranging from robotics to sensor network~\cite{dorri2018multi}, to execute a variety of tasks, e.g., environment sampling~\cite{bellingham2006autonomous}, search and rescue~\cite{xiao2021autonomous, xiao2017uav, xiao2017visual}, and surveillance~\cite{liu2021team}. While enjoying the benefits of accomplishing a complex task with a team of agents, challenges arise correspondingly. For example, task allocation~\cite{khamis2015multi}, consensus and formation control~\cite{ren2005consensus, olfati2006flocking}, collision-avoidance~\cite{WX-JW:22Arxiv, hart2020using}, and communication and synchronization~\cite{khonji2022multi} are problems that do not exist in single-agent scenarios. 

Researchers have studied varying levels of ``teaming''. Earlier works have focused more on the scalability but with less concern on the inter-agent dependency leading to a simple ``divide and conquer'' approach \cite{liu2021team, oh2015survey}. Some more recent works have considered how heterogeneity improves the team performance, including robot teams with heterogeneous capabilities and/or heterogeneous policies (i.e., specialized roles) that arise in teaming behavior~\cite{zheng2011consensus}. In this paper, we are interested in considering a scenario where the team members may dynamically take different roles to cooperatively perform a task, i.e., through coordination.

\subsection{Multi-Agent Reinforcement Learning}

Multi-Agent Reinforcement Learning (MARL)~\cite{marl-book} is an extension of traditional RL that addresses settings with multiple agents that operate either cooperatively, competitively, or coexist in mixed settings. 
MARL algorithms can generally be categorized based on their learning paradigms---centralized, decentralized, and hybrid approaches~\cite{marl-book}. 
In centralized learning, the training process has access to the full states and actions of all agents, enabling a comprehensive learning framework. In contrast, decentralized learning limits each agent to its local observations, making it more suitable for scenarios where global state information is either not available or not practical to use. 
Based on the approach to solve the learning problem, MARL is also categorized into value-based methods like QMIX~\cite{rashid2018qmix} and VDN~\cite{sunehag2017valuedecomposition}, and policy-based methods such as MAPPO~\cite{yu2022surprising}, MADDPG~\cite{lowe2020multiagent}, and MAA2C~\cite{chu2019multi}. Many of these techniques are extended and adapted from single-agent RL algorithms like Q-Learning~\cite{Watkins1992} and PPO~\cite{schulman2017proximal}, and some integrate both value and policy-based strategies like Actor-Critic methods. 

In terms of using MARL to facilitate agent coordination, approaches like DICG~\cite{li2021deep} and DCG~\cite{böhmer2020deep} use graphs as a tool to represent the interaction topology between agents, but not as the environment iteself. In these settings, each node in the graph represents an agent and edges represent communication or influence pathways between them. Agents learn to coordinate their actions either through explicit message-passing along the graph structures (in the case of DCG) or through implicit coordination that leverages the graph topology (as in DICG).

In cases where the environment can be efficiently represented as a graph~\cite{limbu2023team}, these conventional MARL methods lack specialized mechanisms to exploit the graph topology or the node-edge relationships. 
This research gap is what this work aims to fill. In a centralized manner, we convert the original problem of team coordination on graphs into the form of an MDP to present not only agent states, but also the states of the environment around the agents. We show RL has the potential to solve these problems more efficiently when more agents and larger graphs are of interest, comparing to state-of-the-art classical methods.

\section{PROBLEM FORMULATION}
\label{sec::method}

We first present our original problem formulation of team coordination on graphs with state-dependent edge costs~\cite{limbu2023team}, which, in this work, is converted into a novel MDP formulation with a new state and action space, along with a state transition function. We also design a reward function based on the traversal cost to enable reinforcement learning to efficiently solve this problem in two different settings: learning to solve for a single graph and learning to solve for multiple graphs with the same number of nodes.

\subsection{Team Coordination on Graphs}
A team of $N$ agents travel on a strongly connected Environment Graph (\textsc{eg}) denoted by \(\mathbb{G}=(\nodeset,\edgeset)\), where \(\nodeset\) is a set of nodes, and \(\edgeset\) is a set of edges, \(\edgeset\subset \nodeset\times \nodeset\), from a start node set $\nodeset_0 \subset \nodeset$ to reach a goal node set $\nodeset_g\subset \nodeset$ while minimizing the cost of traversal.
The nominal cost for traversing the edge \(\edge{i}{j}\in \edgeset\) is given as a constant, $\cost{i}{j}$ for $i,j\in \nodeset$, when traveling on that edge without support. On the \textsc{eg}, each edge $\edge{i}{j}$ is associated with a set of support nodes, $\supnodes{i}{j} \subseteq \nodeset$. If this set is non-empty, an agent at $v\in\supnodes{i}{j}$ can provide support for the agent traversing $\edge{i}{j}$ and reduce the nominal traversal cost $\cost{i}{j}$ to $\supmove{i}{j}$. 
The action set for any agent at node $i$ is given as $\actset{i}=\{\{\act{}{i,}{j}\}_{j\in\mathcal{N}_i}, \act{}{}{s}\}$, where $\mathcal{N}_i$ is the neighborhood of $i$, and $\act{}{i,}{j}$ is the action to move to node $j$ given that $j\in\mathcal{N}_i$. The action $\act{}{}{s}$ is the support while inducing an extra support cost $\supcost$. 

For agent $n\in \agentset$, denote its action sequence and node visiting sequence as $\actseq{n}=\{a_i^n\}_{i=0}^{T-1}$ and $\nodeseq{n}=\{v_i^n\}_{i=0}^{T}$ respectively, where $T$ is the total time step to reach the goal, $v_0^n \in \nodeset_0$ and $v_T^n \in \nodeset_g$ are the start and goal node, and $a_i^n \in \actset{v_i^n}$. In general, the cost of each action taken by an agent $n \in \agentset$ at each step $i$ is a function $C^n(\cdot)$ of the positions and actions of all agents at $i$: $\agentcost{n}{i} = C^n(\{v_i^j, a_i^j\}_{j\in\agentset})$. The goal of the team is to find the team action sequences $\{\actseq{j}\}_{j \in \agentset}$ in order to minimize the accumulated cost of the entire team along the entire traversal to the goal: 
\begin{equation}
\min_{\{\actseq{j}\}_{j \in \agentset}} \sum_{n\in\agentset} \sum_{i=0}^{T-1} \agentcost{n}{i}. 
\label{eq:min_cost_function}
\end{equation}

\subsection{MDP Formulation}
We convert our original problem into a MDP formulation with a novel state and action space, along with a state transition and reward function. 

\subsubsection{State Space}
A Markovian state needs to incorporate all necessary information along with the action to determine the next state and current reward. Part of our new state $S$ includes all individual agent positions at each time step $t$, i.e., $\{v_t^n\}_{n \in \agentset}$, which is represented as a one-hot vector for each agent and therefore the joint agent state has dimensionality $|\nodeset| \cdot |\agentset|$, i.e., $P_{|\nodeset| \cdot |\agentset|\times1}\in\mathbb{P}_{|\nodeset| \cdot |\agentset|\times1}$. Our state also needs to consider the \textsc{eg} to be Markovian. Therefore, we include the graph connectivity and supporting mechanism as part of $S$. We use the adjacency matrix of the \textsc{eg} to encode graph connectivity, i.e., a $|\nodeset| \times |\nodeset|$ matrix $\textrm{ADJ}_{|\nodeset| \times |\nodeset|} \in \mathbb{ADJ}_{|\nodeset| \times |\nodeset|}$, with each entry $\textrm{adj}_{i, j}$ denoting the nominal traversal cost between node $i$ and $j$ or set as $\infty$ if the edge $\edge{i}{j}$ does not exist. We also include as part of $S$ the supporting mechanism as a supporting tensor $\textrm{SUP}_{|\nodeset| \times |\nodeset| \times |\nodeset|} \in \mathbb{SUP}_{|\nodeset| \times |\nodeset| \times |\nodeset|}$. Each entry $\textrm{sup}_{i, j, k}$ denotes the reduced traversal cost between node $j$ and $k$, if there is an agent taking support action from node $i$. $\textrm{sup}_{i, j, k}$ remains the nominal cost of $\edge{j}{k}$ if support is not possible from $i$ for $\edge{j}{k}$. Therefore, our new state space is defined as 
\begin{equation}
    \mathbb{S} \coloneqq \mathbb{P}_{|\nodeset| \cdot |\agentset|\times1} \times \mathbb{ADJ}_{|\nodeset| \times |\nodeset|} \times \mathbb{SUP}_{|\nodeset| \times |\nodeset| \times |\nodeset|}. 
    \label{eqn::state}
\end{equation}

\subsubsection{Action Space}
Each agent is able to take the action of moving to any neighboring nodes (including staying at the current node and inducing zero cost) or supporting. To assure the action space for each agent has the same dimensionality across all states in our state space $\mathbb{S}$, we define the action for agent $n$ to be a one-hot vector $a^n_{(|\nodeset|+1) \times 1} \in A_{(|\nodeset|+1) \times 1}$, denoting which node the agent moves to (or stays at the current node). The last dimension of $a^n_{(|\nodeset|+1) \times 1}$ denotes the agent is taking the supporting action. In a centralized manner, the joint action space of the entire team can be defined as 
\begin{equation}
    \mathbb{A} \coloneqq A^1 \times A^2, ..., \times A^N. 
    \label{eqn::action}
\end{equation}
Considering that it is impossible to move from some node to others if there is no edge connecting them, we employ invalid action masking for those cases (details in Sec.~\ref{sec::implementations}).

\subsubsection{Reward Function}
Based on the defined state space (Eqn.~(\ref{eqn::state})) and action space (Eqn.~(\ref{eqn::action})), we define our reward function based on the negative cost induced by the entire team: 
\begin{equation}
    R_t \coloneqq \mathbb{R}(S_t, A_t) =  -\sum_{n\in\agentset} \agentcost{n}{t},
    \label{eqn::reward}
\end{equation}
along with a high reward when all agents reach the goal(s). In order to encourage the team to quickly reach the goal, in addition to this original reward, we also provide reward shaping (details in Sec.~\ref{sec::implementations}).

\subsubsection{State Transition Function}
Since our problem is formulated on graphs, the centralized state transition function $S_{t+1} \sim \mathcal{T}(\cdot | S_t, A_t)$ follows the graph adjacency matrix $\textrm{ADJ}_{|\nodeset| \times |\nodeset|}$, which is also part of the state space. While in our current implementation we simplify the state transition function as a deterministic function, we leave the formulation general enough to account for future nondeterministic cases, e.g., taking the action to move from node $i$ to $j$ has a non-zero probability of staying at $i$ or moving to another node.

\subsubsection{Full MDP}
The full MDP is therefore formulated as a tuple, $(\mathbb{S}, \mathbb{A}, \mathcal{T}, \gamma, \mathbb{R}$), with $\gamma$ as a discount factor which we set to 0.95 in our implementation. The overall object of RL is to learn a policy $\pi: \mathbb{S} \rightarrow \mathbb{A}$ that can be used to select team actions in a centralized manner to maximize the expected cumulative reward over time, i.e., 
\begin{equation}
    J = \mathbb{E}_{(S_t, A_t) \sim \pi}[\sum_{t=0}^\infty \gamma^t R_t]. 
    \label{eqn::return}
\end{equation}

\subsection{Reinforcement Learning for Single and Multiple \textsc{eg}(s)}
In this work, we use different RL algorithms, i.e., Q-Learning~\cite{watkins1992q} and PPO~\cite{schulman2017proximal}, to optimize the expected cumulative reward (Eqn.~(\ref{eqn::return})) in two different settings: learning for a single \textsc{eg} and learning for multiple $N$-node \textsc{eg}s.  

\subsubsection{Single \textsc{eg}}
We start with a simplified version of the state space (Eqn.~(\ref{eqn::state})) and remove the adjacency matrix and supporting tensor, i.e., $\mathbb{S} \coloneqq \mathbb{P}_{|\nodeset| \cdot |\agentset|\times1}$. In this setting, we only aim at learning a policy $\pi$ that works on one \textsc{eg} with a pre-defined supporting mechanism, so that the policy learns what is the optimal team joint action for every team joint state in the simplified state space. Using RL, our goal is to achieve faster solution time, including both RL training and inference, compared to the original \textsc{jsg} method~\cite{limbu2023team}, i.e., \textsc{jsg} construction and shortest-path search time. We also aim at extending to teams with more than two agents, which is the maximal number of agents \textsc{cjsg} can efficiently solve. Furthermore, the \textsc{cjsg} solution time will significantly increase when the ratio between the number of risky edges and the total number of edges is large, because support graph then becomes dense and therefore \textsc{cjsg} essentially becomes \textsc{jsg}. While it is counter-intuitive to expect RL to outperform search-based \textsc{jsg} and \textsc{cjsg}, we hypothesize that RL has the potential to address large-scale problems with many nodes and agents and high risky edge ratio. 

\subsubsection{Multiple \textsc{eg}s}
Second, we also use RL to solve the full MDP. In this setting, a policy is learned to solve any \textsc{eg} with any supporting mechanism given a pre-defined node number $N$. The full state space (Eqn.~(\ref{eqn::state})) can be divided into different disconnected subspaces (imagine no matter what actions the team take, they cannot be teleportated from one \textsc{eg} to another). While the training time for such a policy can be much longer, the learned policy can be reused when a new $N$-node \textsc{eg} with a new supporting mechanism is encountered. For this setting, we compare the RL inference time to solve the problem against their classical counterparts. 

\section{IMPLEMENTATIONS}
\label{sec::implementations}
Based on our MDP problem formation of the original team coordination on graphs problem, we present implementation details on the RL algorithms, reward shaping, and invalid action masking. 

\subsection{RL Implementation}\label{subsec: rl_implementation}
We employ two distinct RL techniques, Q-Learning~\cite{watkins1992q} and Proximal Policy Optimization (PPO)~\cite{schulman2017proximal}, to tackle the multi-agent team coordination problem. To accelerate the learning process, we conduct reward shaping and integrate invalid action masking for both algorithms.

\subsubsection{Q-Learning}
We first use Q-Learning, a value based method, to solve our problem formulated as an MDP. In our centralized problem formulation, a global Q-table is used to maintain the Q-values for all joint states and actions for all agents. Actions are selected by alternating between exploration and exploitation using an $\epsilon$-greedy policy.
Both state transitions and rewards are managed globally. We update our Q-function $Q(s, a)$ based on the Bellman equation:
\begin{equation}\
\small
\begin{split}
Q(s, a) \leftarrow Q(s, a) + \alpha \left[ R^{*}(s, a) + \gamma \max_{a'} Q(s', a') - Q(s, a) \right],
\end{split}
\end{equation}
where $R^{*}(s, a)$ represents a shaped reward detailed in Sec.~\ref{subsec: reward_shape}, $\alpha$ is the learning rate, and $\gamma$ is the discount factor.

Q-Learning encounters scalability issues when facing larger problems involving more agents and nodes, as the extensive Q-table does not fit within memory constraints. As the problem size increases, Q-Learning's performance begins to degrade and eventually fails to solve very large problems.

\subsubsection{PPO}
To address such scalability challenges, we also implement an on-policy method, PPO. 
Like its Q-Learning counterpart, our PPO model is also centralized, but it employs a multi-discrete, one-hot encoded representation for the state space defined in Eqn.~(\ref{eqn::state}). Actions for multiple agents are taken from a multi-discrete action space, facilitating simultaneous actions of all agents as defined in Eqn.~(\ref{eqn::action}). 
An important component of our PPO implementation is the modified clipped loss function, formalized as follows:
        
\begin{equation}
\begin{aligned}
L_{\text{clip}} = \mathbb{E}[ \min \bigg( \frac{\pi(a|s)}{\pi_{\text{old}}(a|s)} & A^{*}(s, a), \\
\text{clip}\bigg(\frac{\pi(a|s)}{\pi_{\text{old}}(a|s)}, 1-\epsilon, 1+\epsilon \bigg) & A^{*}(s, a) \bigg) ],
\end{aligned}
\end{equation}
where $ \frac{\pi(a|s)}{\pi_{\text{old}}(a|s)} $ is the likelihood ratio that compares how likely the current policy $ \pi(a|s) $ scores an action $ a $ given a state $ s $ to how likely the old policy $ \pi_{\text{old}}(a|s) $ scores the same action, $ \text{clip}(\cdot, 1-\epsilon, 1+\epsilon) $ is a clipping function that limits the value within the range $ [1-\epsilon, 1+\epsilon] $,
and the advantage function $ A^{*}(s, a) $ incorporates the shaped reward $R^{*}(s, a)$ (Sec.~\ref{subsec: reward_shape}) to efficiently guide policy optimization.

\subsection{Reward Shaping}\label{subsec: reward_shape}
The original problem formulation for team coordination on graphs is to minimize the entire team's traversal cost on the way to the goal(s) as shown in Eqn.~(\ref{eq:min_cost_function}), which needs to be converted into a reward function and a return as shown in Eqn.~\eqref{eqn::reward} and \eqref{eqn::return} for RL respectively. We also shape our reward function to encourage the agents to explore different coordination options to potentially reduce cost and improve reward. Our reward function includes the following terms: 

\paragraph{Goal Reward}
\begin{equation}
r_g = \left\{
    \begin{array}{cc}
        +10, & \mbox{if } \text{all agents arrive at goal(s)},  \\
        -0.01, & \mbox{otherwise.}
    \end{array}
    \nonumber
\right.
\end{equation}
This reward term assigns +10 when all agents reaching their goal node(s) on the graph. The episode will then be terminated. To encourage moving to the goal fast, every other time step will be penalized by a -0.01 reward. 

\paragraph{Movement Reward}
\begin{equation}
r_m = -\sum_{n\in\agentset} \agentcost{n}{t} = -\sum_{n\in\agentset} C^n(\{v_t^j, a_t^j\}_{j\in\agentset}). 
\nonumber
\end{equation}
This reward is computed by summing up and negating the individual costs incurred by all the agents taking one action at one time step $t$, including incurring regular, risky, or reduced cost when traversing the edge, providing support, and do nothing. In our experiments, normal edge traversal cost is around 1, while risky edges cost around 2, which can be reduced to around 0.5 when being supported. 

\paragraph{Coordination Reward}
\begin{equation}
r_c = \alpha \times \textrm{CC} - \beta \times \text{RC}, 
\nonumber
\end{equation}
where CC and RC denotes the total count of coordination (providing support to reduce risk) and the total count of an agent traversing a risky edge without support with $\alpha=2$ and $\beta=5$. This reward is additional to the original cost function in Eqn.~(\ref{eq:min_cost_function}), since coordination is not absolutely necessary and traversing risky edges without support does not have to be avoided, if the total traversal cost can be kept low. But we find that in practice it helps to encourage the agents to explore different coordination strategies to eventually reduce traversal cost or improve accumulated reward overall. 

\paragraph{Final Reward}
The final reward is therefore a weighted sum of all aforementioned reward terms: 
\begin{equation}
r = w_1 r_g + w_2 r_m + w_3 r_c, 
\nonumber
\end{equation}
where $w_1$, $w_2$, and $w_3$ are weights for the reward terms and set to 1, 1, and 0.2, based on empirical results.

\subsection{Invalid Action Masking}\label{subsec: action_mask}

In our MDP formulation, we represent the individual agent's action as a one-hot vector, indicating the node the agent moves to. However, such an action definition inherently includes numerous invalid actions for nodes that are not directly connected to the agent's current position. Given that valid actions are defined as moving to neighboring nodes or supporting, we implement invalid action masking~\cite{Huang_2022} to restrict permissible actions based on the graph structure at different nodes. This approach excludes invalid actions from consideration during the decision-making process.

In the case of Q-Learning, invalid actions are masked and only valid actions are considered when exploiting Q-values and when exploring the action space randomly.
We design a mask function $m = M(s, a) \in \{0, 1\}$, returning $1$ if action  $a$ is valid and $0$ if invalid in a given state $s$. 
Note that in a centralized manner, any invalid action from any agent in the team will cause the total action to be invalid. 
During exploration, an action $a$ is selected according to an $\epsilon$-greedy policy based on the Q-values and the mask function $ M(s, a) $:

\begin{equation}
a = 
\begin{cases} 
\argmax_{a'}  (Q(s, a')\times M(s, a')),~\text{with prob. } 1 - \epsilon, \\
\text{random action $a'$, s.t. } M(s, a') = 1,~\text{with prob. } \epsilon.
\end{cases}
\end{equation}
Here, the first case picks the valid action that maximizes the masked Q-value with probability $1 - \epsilon$, while the second case picks a random valid action (according to $M(s, a')$) with probability $\epsilon$.

In PPO, the policy’s action selection mechanism is refined by modifying action log probabilities to account for the validity of actions in the current state. This is achieved by applying a mask to action probabilities:

\begin{equation}
\pi(a \mid s) = \frac{\exp(\log \pi(a \mid s))}{\sum_{a', \text{ s.t. } M(s, a')=1} \exp(\log \pi(a' \mid s))}, 
\end{equation}

Here, action probabilities, $\pi(a \mid s)$, are recalculated to consider only valid actions (where $M(s, a')=1$) as determined by the mask function $M(s, a)$ effectively eliminating the chance of selecting invalid actions.

We observe that reward shaping (Sec.~\ref{subsec: reward_shape}) and invalid action masking enhance sample efficiency, reduce training time and accelerate convergence for both Q-learning and PPO.

\section{RESULTS}
\label{sec::results}

\begin{table*}[t]
\centering
\caption{Solution time for 2, 3, and 4 agents in JSG, Q-Learning and PPO respectively.}
\label{tab::time_comparison}
\small
\setlength{\tabcolsep}{2.5pt} 
\begin{tabular}{ccccccccccc}
\toprule
Graph & Nodes & \multicolumn{3}{c}{2 Agents} & \multicolumn{3}{c}{3 Agents} & \multicolumn{3}{c}{4 Agents} \\
\cmidrule(rl){3-5} \cmidrule(rl){6-8} \cmidrule(rl){9-11} & & JSG & Q-Learning & PPO & JSG & Q-Learning & PPO & JSG & Q-Learning & PPO \\
\midrule
\multirow{5}{*}{Sparse} 
& 5  & \textbf{0.001} & 1.228 & 58.39 & \textbf{0.037} & 2.863 & 83.66 & \textbf{1.093} & 9.978 & 88.74 \\
& 10 & \textbf{0.014} & 3.654 & 81.39 & \textbf{1.494} & 10.35 & 226.2 & 157.7 & \textbf{102.7} & 355.8 \\
& 15 & \textbf{0.057} & 5.922 & 201.02 & \textbf{14.88} & 27.88 & 326.4 & 3652 & -- & \textbf{962.2} \\
& 20 & \textbf{0.172} & 13.86 & 560.3 & 80.16 & \textbf{45.31} & 701.5 & -- & -- & \textbf{1045} \\
& 25 & \textbf{0.394} & -- & 730.5 & \textbf{281.0} & -- & 1432 & -- & -- & -- \\
\midrule
\multirow{5}{*}{Moderate} 
& 5  & \textbf{0.002} & 0.293 & 56.97 & \textbf{0.052} & 3.469 & 74.03 & \textbf{1.689} & 14.34 & 88.21 \\
& 10 & \textbf{0.022} & 2.362 & 66.17 & \textbf{3.007} & 20.36 & 146.3 & 600.5 & 751.0 & \textbf{352.2} \\
& 15 & \textbf{0.088} & 2.389 & 189.6 & 25.49 & \textbf{22.79} & 317.7 & 9492 & -- & \textbf{949.4} \\
& 20 & \textbf{0.277} & 3.587 & 531.0 & 160.04 & \textbf{58.26} & 683.3 & -- & -- & \textbf{1032} \\
& 25 & \textbf{0.641} & 5.720 & 677.5 & 571.1 & \textbf{181.8} & 1372 & -- & -- & -- \\
\midrule
\multirow{5}{*}{Dense} 
& 5  & \textbf{0.002} & 0.874 & 57.32 & \textbf{0.072} & 1.855 & 72.44 & \textbf{0.072} & 6.921 & 89.71 \\
& 10 & \textbf{0.035} & 1.963 & 64.35 & \textbf{7.927} & 15.11 & 142.3 & 4312 & 696.9 & \textbf{344.8} \\
& 15 & \textbf{0.109} & 6.671 & 186.4 & \textbf{39.49} & 129.1 & 317.7 & 46455 & -- & \textbf{944.1} \\
& 20 & \textbf{0.433} & 2.616 & 646.2 & 481.4 & \textbf{65.22} & 677.5 & -- & -- & \textbf{1018} \\
& 25 & \textbf{0.915} & 5.192 & 700.9 & 1660 & 7821 & \textbf{1349} & -- & -- & --\\
\bottomrule
\end{tabular}
\vspace{-20pt}
\end{table*}

With our new problem formulation and implementation, we conduct extensive experiments to study how RL can efficiently enable team coordination on graphs with risky edges. We present experiment results of using RL to solve both single and multiple \textsc{eg}(s).

\subsection{RL for Single \textsc{eg}}
The first set of experiments is to use RL to solve one single \textsc{eg}. Despite that the classical \textsc{jsg} approach can be efficient in solving simple coordination problems on small graphs with a small number of agents, we hypothesize that RL has the potential to solve complex problems faster. Note that since \textsc{cjsg} can only address problems with up to two agents, and our goal is to extend to more than two, we do not include \textsc{cjsg} in our comparison. 

Specifically, we experiment with 5, 10, 15, 20, and 25 nodes with 2, 3, and 4 agents in three types of graph connectivity, i.e., sparse, moderate, and dense, using \textsc{jsg}, Q-Learning, and PPO. We randomly create 15 \textsc{eg}s as our test set. All experiment results are presented in Tab.~\ref{tab::time_comparison}. For both RL approaches, training is terminated when the cumulative reward no longer changes more than 0.2 for 500 steps. 

The results in Tab.~\ref{tab::time_comparison} shows that \textsc{jsg} outperforms both Q-Learning and PPO in all two-agent cases, indicating that the cost to construct a \textsc{jsg} for two agents and search on such a graph is minimal even with up to 25 nodes. However, RL's superiority starts to show when the number of agents and nodes and graph connectivity start to increase. For three agents, Q-Learning starts to outperform \textsc{jsg} in 20-node sparse graphs, while it completely overtakes \textsc{jsg} for moderate graphs with more than 15 nodes. Q-Learning and PPO outperform \textsc{jsg} on 20-node and 25-node dense graphs respectively. In most cases with four agents, PPO (and Q-Learning) is the fastest, except for small 5-node graphs. Notice that ``--'' denotes that the algorithm fails to find a solution, i.e., \textsc{jsg} runs out of memory during graph construction, Q-Learning's Q-table becomes intractably large, or PPO does not converge. All three methods fail to produce a solution for the most difficult case at the lower right of Tab.~\ref{tab::time_comparison}, i.e., four agents on 25-node dense graphs, while PPO is the only one that can solve for four agents on 20-node dense graphs.

\begin{figure*}
    \centering
    \includegraphics[width=0.32\textwidth]{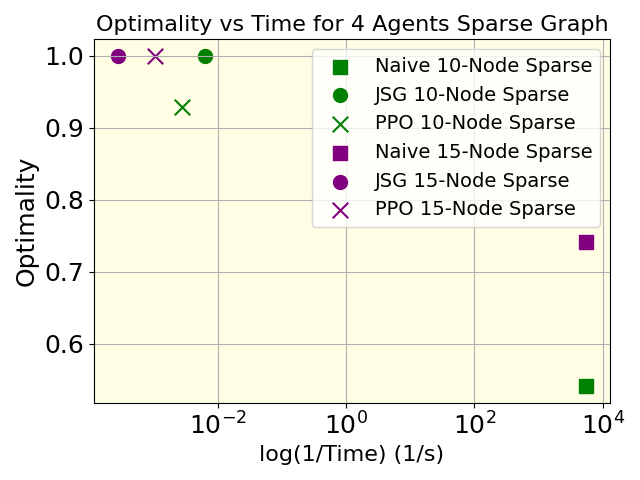}
    \includegraphics[width=0.32\textwidth]{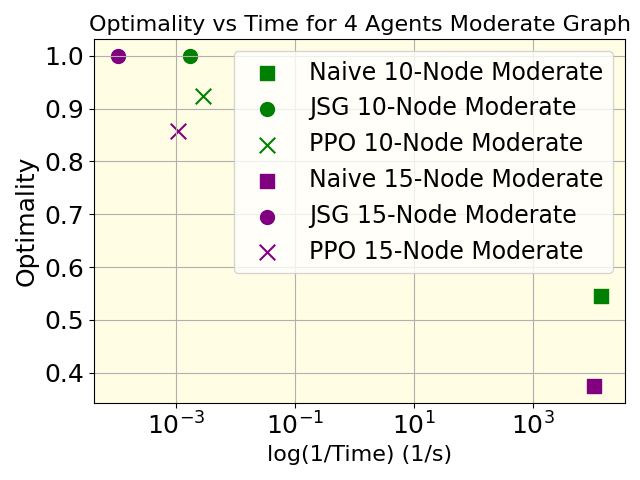}
    \includegraphics[width=0.32\textwidth]{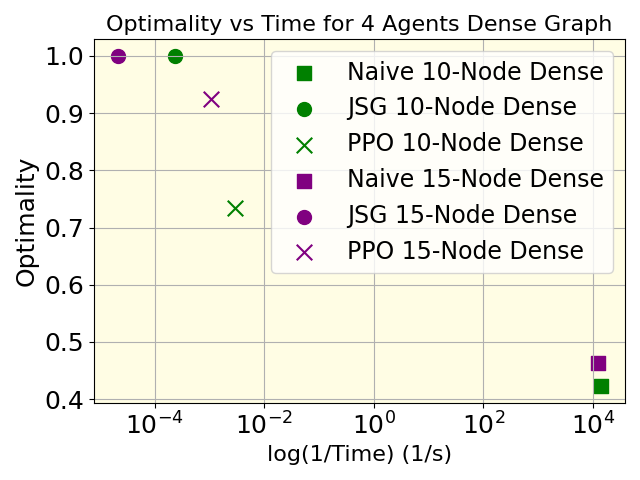}
    \caption{Single \textsc{eg}: Optimality vs. Time plots for four agents on sparse, moderate, and dense graphs.}
    \label{fig::pareto_1st_half}
    \vspace{-10pt}
\end{figure*}

For solving one \textsc{eg}, while RL cannot guarantee optimality, we observe that in most cases RL can achieve optimal solutions, and near-optimal ones in others. We use the four agents in 10- and 15-node graphs as an example and present the optimality vs. time plots in Fig.~\ref{fig::pareto_1st_half}. We also implement a naive approach, in which all agents do not seek coordination, but just move towards the goal with the minimal cost path. The naive approach is very efficient in terms of time, since it only needs to call a shortest-path algorithm once and the solutions for all agents remain the same, but without coordination the naive approach will incur a large cost. On the other hand, \textsc{jsg} is provably optimal~\cite{limbu2023team}, so we use the ratio between a solution's cost and the optimal \textsc{jsg} cost as an indication of solution optimality. In Fig.~\ref{fig::pareto_1st_half}, the horizontal axis is the reciprocal of the solution time in log scale, while the vertical axis is the optimality value between 0 and 1. Naive approach is expected to appear in the lower right corner by achieving very short solution  time but very low path quality, whereas \textsc{jsg} should be in the upper left corner with optimality value 1 and a very long solution time. 
For four agents on sparse 10-node graphs, PPO does not produce any advantage over \textsc{jsg}, by achieving lower optimality with longer time. But for sparse 15-node graphs, PPO outperforms \textsc{jsg} by achieving optimality with better time efficiency. For moderate 10- and 15-node graphs, PPO achieves a middle ground in terms of both optimality and time between \textsc{jsg} and the naive approach. We observe the same optimality-time trend in dense graphs of 10 and 15 nodes. In all scenarios, the worst optimality ratio PPO can achieve is more than 70\% of \textsc{jsg}'s absolute optimality value, but mostly with better time efficiency than \textsc{jsg}.

\begin{figure*}
    \centering
    \includegraphics[width=0.32\textwidth]{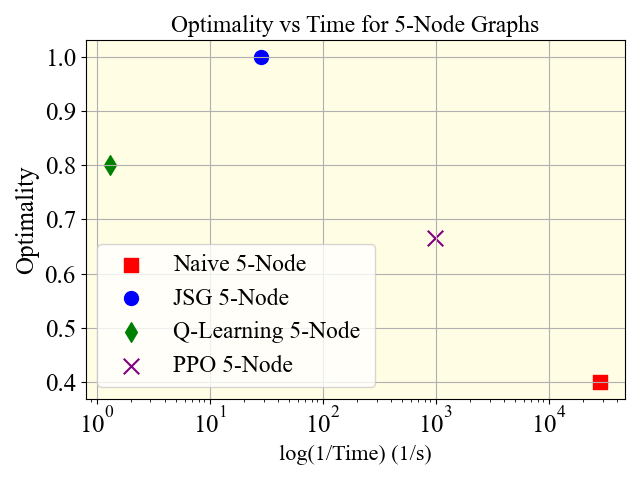}
    \includegraphics[width=0.32\textwidth]{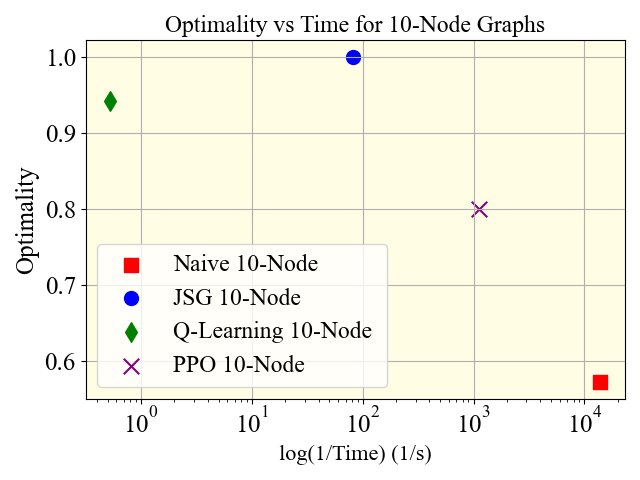}
    \caption{Multiple \textsc{eg}: Optimality vs. Time plots for two agents on any 5- or 10-node graphs.}
    \label{fig::pareto_2nd_half}
    \vspace{-20pt}
\end{figure*}

\subsection{RL for Multiple \textsc{eg}}

The second set of experiments  is to use RL to solve any $N$-node graphs. This is a very difficult task considering the variations in graph connectivities and supporting mechanisms with a large number of nodes. Therefore, we only limit our experiments within up to 10 nodes and 2 agents and leave extending to more complex problems to future work, potentially with decentralized approaches. Notice that when RL converges to a good policy, despite the long time it may require, this policy can then be used as an available tool to solve any team coordination problem on any $N$-node graph in the future. In the second set of experiments, training usually takes hours, which is considered as a one-time cost. Once trained, we compare RL's inference time and optimality of solving any $N$-node graph with \textsc{jsg} and the naive approach, as shown in Fig.~\ref{fig::pareto_2nd_half}. Q-Learning does not scale well to this challenging problem by underperforming \textsc{jsg} in terms of both optimality and time , while PPO can find the middle ground in terms of optimality and time between \textsc{jsg} and the naive approach. The PPO results (magenta crosses) in both graphs indicate that for any 5-node and 10-node graph, on average PPO can solve it with 70\% and 80\% optimality and half of the time compared to \textsc{jsg} respectively.

\section{CONCLUSIONS}
\label{sec::conclusions}

We study RL techniques to enable team coordination behaviors in graph environments with support actions among teammates to reduce edge traversal costs in a centralized manner. By converting the original team coordination on graphs with risky edges problem into a novel MDP formulation, we are able to apply RL to solve it. Our proposed state space is able to capture not only robot positions, but also graph connectivities and supporting mechanisms. 
Our experiment results indicate that while classical approaches can solve simple problems with smaller number of nodes and agents very efficiently with optimality guarantee, RL has the potential to outperform classical approaches in larger graphs with more agents. 
However, there is still room to improve with respect to graph scale, team size, optimality, and efficiency with better state and action space and shaping reward design. Another promising direction is to move towards the decentralized regime to keep improving on scalability with provable and bounded reduction on optimality.

\bibliographystyle{ieeetr}
\bibliography{bib}
\end{document}